\documentclass[letterpaper]{article} 

\pdfoutput = 1

\usepackage{aaai24}  
\usepackage{times}  
\usepackage{helvet}  
\usepackage{courier}  
\usepackage[hyphens]{url}  
\usepackage{graphicx} 
\usepackage{natbib}  
\usepackage{caption} 
\usepackage{algorithm}
\usepackage{algorithmic}
\usepackage{tcolorbox}
\usepackage[inline]{enumitem}
\usepackage{booktabs}
\usepackage{tablefootnote}
\usepackage[flushleft]{threeparttable}
\usepackage{amsmath}
\usepackage{xcolor}
\usepackage{soul}

\newcommand{\sg}[1]{\textcolor{purple}{#1}} 
\newcommand{\nk}[1]{\textcolor{cyan}{#1}} 

\usepackage{newfloat}
\usepackage{listings}
\DeclareCaptionStyle{ruled}{labelfont=normalfont,labelsep=colon,strut=off} 
\floatstyle{ruled}
\newfloat{listing}{tb}{lst}{}
\floatname{listing}{Listing}
\title{Exploring Failure Cases in Multimodal Reasoning About Physical Dynamics}
\author{
    Sadaf Ghaffari, Nikhil Krishnaswamy\\\
}
\affiliations{
    Colorado State University, Fort Collins, CO USA \\
    \{sadafgh,nkrishna\}@colostate.edu
}

\begin{document}

\maketitle

\begin{abstract}
In this paper, we present an exploration of LLMs' abilities to problem solve with physical reasoning in situated environments. We construct a simple simulated environment and demonstrate examples of where, in a zero-shot setting, both text and multimodal LLMs display atomic world knowledge about various objects but fail to compose this knowledge in correct solutions for an object manipulation and placement task. We also use BLIP, a vision-language model trained with more sophisticated cross-modal attention, to identify cases relevant to object physical properties that that model fails to ground. Finally, we present a procedure for discovering the relevant properties of objects in the environment and propose a method to distill this knowledge back into the LLM.
\end{abstract}

\section{Introduction}

\begin{figure}[h!]
    \begin{tcolorbox}[title=\textsc{\small Example Prompt}]
    \small
    You need to get to the top of a platform that is 2 meters high.  The highest you can jump is 1 meter.  You have two blue cubes that are both 1 meter long on all sides, a blue sphere that is 1 meter in diameter, and a blue cylinder whose major axis is 1 meter long.  How can you get to the top of the platform?
    \end{tcolorbox} 

    \begin{tcolorbox}[title=\textsc{\small Example Response (LLaMA 2-7B)}]
    \small
    You can use the cube on top of the sphere to get to the desired height. Here's how:
    \begin{enumerate}
        \item Place the cube on top of the sphere. This will give you a total height of 1.5 meters (1 meter for the cube + 0.5 meters for the sphere).
        \item Place the cylinder on top of the cube. This will give you a total height of 2 meters (1 meter for the cube + 1 meter for the cylinder).
    \end{enumerate}
    \end{tcolorbox}
    \centering
    \includegraphics[width=.3\columnwidth]{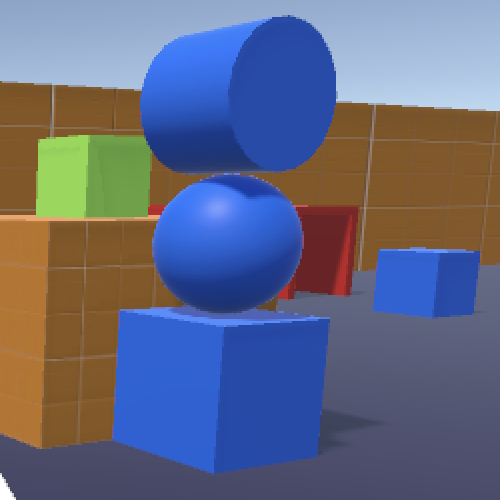}
    \includegraphics[width=.3\columnwidth]{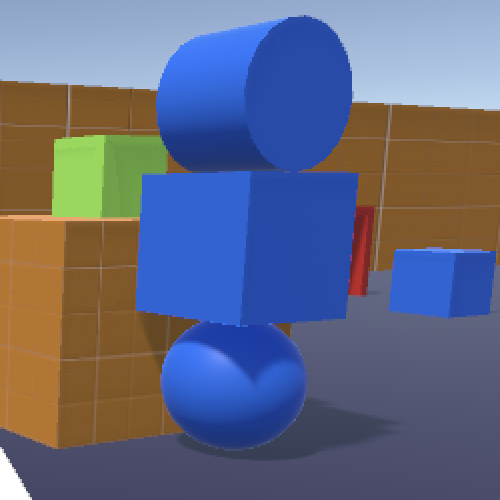}
    \includegraphics[width=.3\columnwidth]{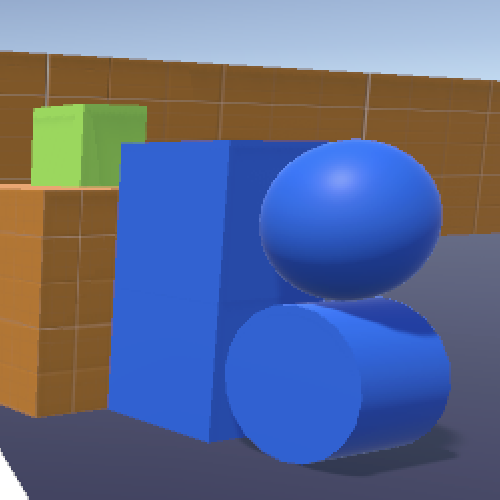}
    \caption{Example of physical reasoning prompt and response, and operationalizations of solutions as given by ChatGPT [L], LLaMA 2-7B~\cite{touvron2023llama} [C], and LLaVA ~\cite{liu2023visual} [R].}
    \label{fig:ex}
\end{figure}

Large Language Models (LLMs), exemplified by modern generative systems like GPT-4, display impressive performance on benchmark tasks~\cite{qin2023chatgpt}. However, these models continue to struggle with questions involving physical reasoning~\cite{goertzel2023generative,wang2023newton}, and demonstrate an apparent lack of ``understanding'' of common-sense physical properties and natural laws. Reasoning about properties of simple objects (such as balls being round) appeared to be weak in encoder-style language models such as BERT~\cite{devlin-etal-2019-bert}. This could be partially attributed to such knowledge being so common as to be largely uninformative, and sparse in training corpora~\cite{krishnaswamy2022affordance}. Newer and larger generative language models appear not to have this problem~\cite{ghaffari2023grounding}, with the autoregressive mechanism able to generate correct output about object properties and affordances. However, this knowledge appears to be atomic, unable to be composed into larger-scale reasoning. Solutions to simple problems that are intuitive to humans appear to not be considered by LLMs despite the level of world knowledge shown about individual items. An example that we will return to throughout this paper is the prompt in Fig.~\ref{fig:ex} and solutions thereto.

None of these solutions account for the properties of objects and physical laws like inertia and gravity. We argue that this is due to a lack of grounding mechanisms to physical environments and access to non-linguistic modalities beyond imagery.

In this paper, we explore the current abilities of LLMs to perform reasoning about simple objects in dynamic environments. We construct a simulation environment in which to investigate these problems, and develop methods to evaluate LLMs' solutions to a reasoning problem in terms of correctness of object selection and physical feasibility. We also examine the cross-modal grounding between images from our environment and key terms in the reasoning task using a cross-attention architecture, and expose a number of shortcomings in this ability relative to the key concepts underlying our example problem. We develop a procedure by which an agent may exploit failure in the zero-shot outputs of LLMs as a trigger to investigate alternative solutions to the problem using object interactions and previously-encoded knowledge of the object semantics. Finally, we propose a method through which knowledge gained from the object interactions can be distilled back into the LLM, and avenues for future research.

\section{Related Work}

Recent explorations of generative LLMs have exposed certain classes of problems that they struggle with.

\paragraph{Linguistic and semantic problems}

\citet{chaturvedi2022analyzing} show that LLM failures in the semantics of tasks like question answering result from features of the training data and regime. \citet{asher2023limits} examine the capabilities of LLMs (broadly defined to include both BERT-style and GPT-style models) in semantic problems with a primary focus on quantification.  They find that LLMs struggle to learn semantic properties of entailment and consistency, according to formal definitions, and do not generalize beyond the first level of the Borel Hierarchy.  Additionally they find that generative models display inconsistency of outputs in this domain, even when temperature is set to 0.

\citet{kocon2023chatgpt} evaluated both ChatGPT and GPT-4 on 25 diverse NLP/NLI tasks and showed that both models approached but did not exceed state of the art performance on mutliple tasks, when compared to performance of models specialized for individual or subsets of tasks.

\paragraph{Mathematical and logic problems}

In what is likely the first independent empirical evaluation of ChatGPT, \citet{shakarian2023independent} showed that its performance on math word problems (MWP) was strongly conditioned on a requirement to show its work, {\it a la} chain-of-thought (CoT) reasoning.  However, they also demonstrated that, even given this condition, ChatGPT's failure rate increases linearly relative to the number of addition and subtraction operations.  We also demonstrate failures of LLMs in multistep reasoning problems, albeit in a different domain.

\citet{chen2023chatgpt} evaluated the performance of ChatGPT (as powered by GPT-3.5 and GPT-4) at different points in time on various problem types, including mathematical and multi-hop reasoning problems.  They discovered unpredictable performance and found that the behavior of the same closed service can change substantially over time as models are updated. Increased performance on one type of tasks may have unpredictable effects on other types.

\citet{shanahan2022talking} maintains a division between the statistical functioning of LLMs and logical structures and methods underlying human reasoning, and mentions how prompt engineering techniques like CoT reasoning effectively reframe the LLM's problem from a logical one to a statistical one, asking the model to sample from a distribution rather than make an {\it a priori} inference.

\citet{goertzel2023generative} argues that shortcomings in LLMs stem from a lack of cognitive architectures underlying the system, and that incremental (i.e., scale- and data-based) approaches are unlikely to be a path from current LLMs to artificial general intelligence (AGI). Our purview in this paper is not this problem specifically, though we do address how an underlying knowledge base mediated by a simulation can allow a neural system to solve a problem a zero-shot LLM apparently cannot, and propose a method for transferring some of this knowledge into an LLM.

\paragraph{Physical reasoning problems}

LLMs display often proficient performance on questions involving knowledge of physics~\cite{polverini2023understanding}, but when it comes to applying common-sense physical reasoning in specific situations, even the most advanced multimodal models (e.g., Say-Can~\cite{ahn2022can} or PALM-E~\cite{driess2023palm}) are highly sensitive to changes in the environment and input. Executing tasks in new situations usually involves crafting additional low-level behaviors for the new environment.

The domain in which we conduct our experiments is immediately reminiscent of pick-and-place tasks~\cite{lobbezoo2021reinforcement}, wherein an agent, usually a robotic arm, must select a target object and place the object at a specified location and orientation.  This has been a popular problem for reinforcement learning (RL) approaches~\cite{li2020towards} and has also been adapted to train models to learn about physical intuitions using tasks such as fall prediction~\cite{lerer2016learning}.  With the advent of generative LLMs, research in the object placement domain has explored how to use LLMs to serve as the back-end planner in situated environments~\cite{ahn2022can,driess2023palm}. In these cases, a standard constraint of pick and place tasks is still in force: the locations of objects must be known or deterministic. In addition, grounding objects to the linguistic input follows a standard approach of joint training over image and text distributions, with no multimodal signal besides pixels, and only integrates language with static images, not situated continuous environments.


\section{Physical Reasoning in LLMs}

LLMs' difficulty with physical reasoning is manifest in Fig.~\ref{fig:ex}, where solutions to the problem result in infeasible and unsafe configurations. The LLM is not grounded to the environmental dynamics.  To further explore the extent to which an LLM does or does not generate correct outputs regarding relations and physical dynamics between objects, we created a simple scene against which to evaluate LLM outputs in this domain, along with a set of controls on prompts and image inputs. We compare outputs from ChatGPT, LLaMA 2~\cite{touvron2023llama}, and LLaVA, an open-weight model that integrates CLIP ViT-L/14~\cite{radford2021learning} with a LLaMA--based generative language model.\footnote{We used the 7 billion parameter version of LLaVA-1.5 in these experiments.}

\subsection{Methodology}

\begin{figure}
    \centering
    \includegraphics[width=.5\columnwidth]{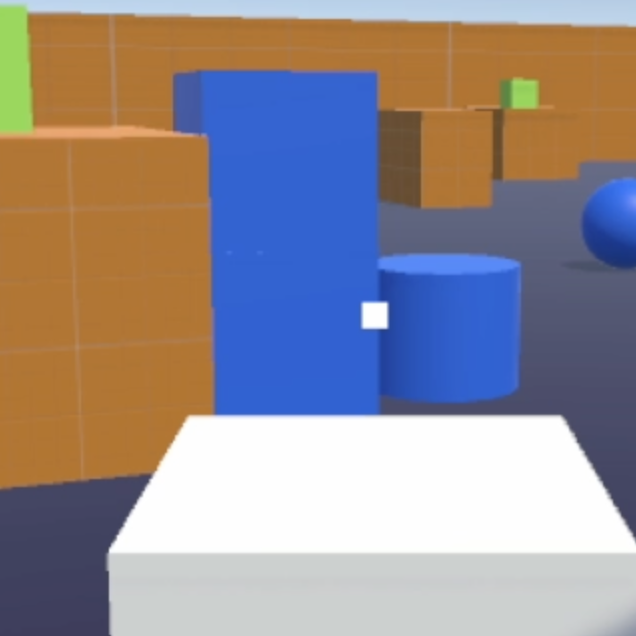}
    \caption{A feasible, physically stable solution to the platform reach problem, that uses the cylinder in an orientation that exploits the stability affordances of its flat ends.}
    \label{fig:correct-solution}
\end{figure}

Following previous research~\cite{ghaffari2022detecting,ghaffari2023grounding}, we focused on a simple set of geometric objects inspired by children's toys, and on the key concepts of flatness and roundness that determine the stability of structures constructed with these objects. We constructed a simple scene in the VoxWorld platform~\cite{krishnaswamy2022voxworld} that reflected the scenario alluded to in Fig.~\ref{fig:ex}. The scene contained 5 interactable objects, each roughly one meter tall/wide---two cubes, two cylinders, and a sphere---and 5 two-meter high platforms. In this paper, we considered only the task of moving objects and stacking them in a configutation to achieve a specified goal.


An agent (represented as a white cube) must navigate the scene and collect objects at the top of each platform. The agent could pick up, carry, rotate, place, and jump on top of any of the interactable objects, with a jump height limited to one meter. The VoxWorld platform for interactive agents takes in linguistic input and operationalizes it in the 3D world by converting the object motions described therein into movements within the environment, which were then executed in real time. Instructions in the output that did not describe moving or placing objects were ignored. Underspecified instructions were handled according to \citet{krishnaswamy2017monte}, by randomly sampling locations that satisfied logically inferrable constraints.

The prompts we explored are variants of that given in Fig.~\ref{fig:ex}, with changes to control for the output format and type of reasoning we wanted to push the LLM toward. We provided the prompt alone to ChatGPT and LLaMA 2, and to LLaVA along with an image of the scene, including the objects mentioned in the prompt. Prompts were slightly modified for the given model, such as by using ``You are in the room shown in the image" with multimodal LLaVA to draw attention to the visual input. We used default temperature values for LLaMA 2 and LLaVA (0.6 and 0.2, respectively).

Use of a fixed problem allows us to assess LLMs' output relative to a limited set of correct solutions, all of which involve building a staircase-like structure of the kind shown in Fig.~\ref{fig:correct-solution}. We can therefore quickly calculate an intersection-over-union (IoU) of the two sets: the objects that the LLMs included in its solution, and the objects in the correct solution (either two cubes and a cylinder or two cylinders and a cube). This calculation concerns only which objects were mentioned by the LLM in the generated solution, and not how they were placed. Placement is evaluated using the \textit{stable} metric as discussed below.  The outputs were then operationalized in VoxWorld and the simulation run. After applying environmental physics, we scored the stability of the structure as the percentage of objects that remain in place. Therefore a correct solution should have a stability of 100\% (all objects remain in place) and an IoU of 1.0 (all correct objects and only correct objects were selected). 
These two metrics indicate how right or wrong an LLM's response is in the environment presented. While they may be somewhat abstract, evaluating at this high level allows us to examine the physical feasibility and correctness of the solution provided, without noise introduced in the specific word-level output by known problematic phenomena of LLMs such as hallucinations or problems with counting or size-based reasoning. They are intentionally tuned to the stacking task and provide a template of how to evaluate physical reasoning in tasks such as this, i.e., measuring both the object selection and key task-relevant concept.

\subsection{Open-world Simulation}

In our first setting, the environment was left relatively unconstrained. Objects remained in the locations in which they spawned, and initially various distractors remained in place (e.g., the collectable items on top of the platforms and the player, which were all rendered as cubes, all remained visible). The prompt also terminated in a relatively open-ended question {\it a la} Fig.~\ref{fig:ex}, resulting in a \textit{free-text response}. A variant on the prompt constrained the response to be a \textit{single sentence} instead of free text (by providing the additional command ``Provide your response in one sentence''). 

Text-only models (ChatGPT, LLaMA 2) were given the prompt alone {\it a la} Fig.~\ref{fig:ex}. LLaVA was also provided the complete image of the scene, including the player (white cube, which can be made invisible) and collectible objects (green cubes).
Table~\ref{tab:open-world} shows scores for LLM outputs.

\begin{table}
    \centering
    \begin{tabular}{lccccc}
    \toprule
    & \multicolumn{2}{c}{Free-text} && \multicolumn{2}{c}{1-sentence} \\
    \cmidrule{2-3} \cmidrule{5-6}
                  & Stable   & IoU       && Stable   & IoU     \\
    \midrule
    {\bf ChatGPT} & .33         & .50       && .67         & .75        \\
    {\bf LLaMA 2-7B} & .00         & .50       && .50         & .50       \\
    {\bf LLaMA 2-70B} & .25         & .75       && 1.00$^\dag$         & .33         \\
    {\bf LLaVA}   & .50         & .75       && .50         & .75         \\
    \midrule
    & \multicolumn{5}{c}{No player} \\
    \midrule
    & \multicolumn{2}{c}{Free-text} && \multicolumn{2}{c}{1-sentence} \\
    \cmidrule{2-3} \cmidrule{5-6}
                  & Stable   & IoU   && Stable   & IoU  \\
    \midrule
    {\bf LLaVA}   & .25      & .75   && .67         & .75 \\
    \bottomrule
    \end{tabular}
    \caption{LLM scores from open-world simulation. The indicated LLaMA 2 models are versions optimized for dialogue (LLaMA 2-chat-*). {\it No player} denotes variant image inputs (hence only provided to LLaVA), in which the player was not rendered. $^\dag$LLaMA 2-chat-70B's response in this condition only mentioned one object, which remains stable by default in the simulation.}
    \label{tab:open-world}
\end{table}

Sample responses such as ``{\it Place one cube on the ground, stand on it, then stack the other cube and the sphere on top of each other to reach the top of the 2-meter high platform}'' (ChatGPT) and ``{\it The blue cylinder, the blue sphere, and the blue cube on the side that is one meter long}'' (LLaVA) demonstrate the same problems as seen in Fig.~\ref{fig:ex}: a lack of consideration for the effect of object properties on environmental dynamics (such as what happens when trying to stack another object on a sphere), and also a failure to consider all information in the prompt (such as the fact that both cubes are of the same size).

\subsection{Controlled Environment}

\begin{figure*}[h!]
    \centering
    \includegraphics[width=.3\textwidth]{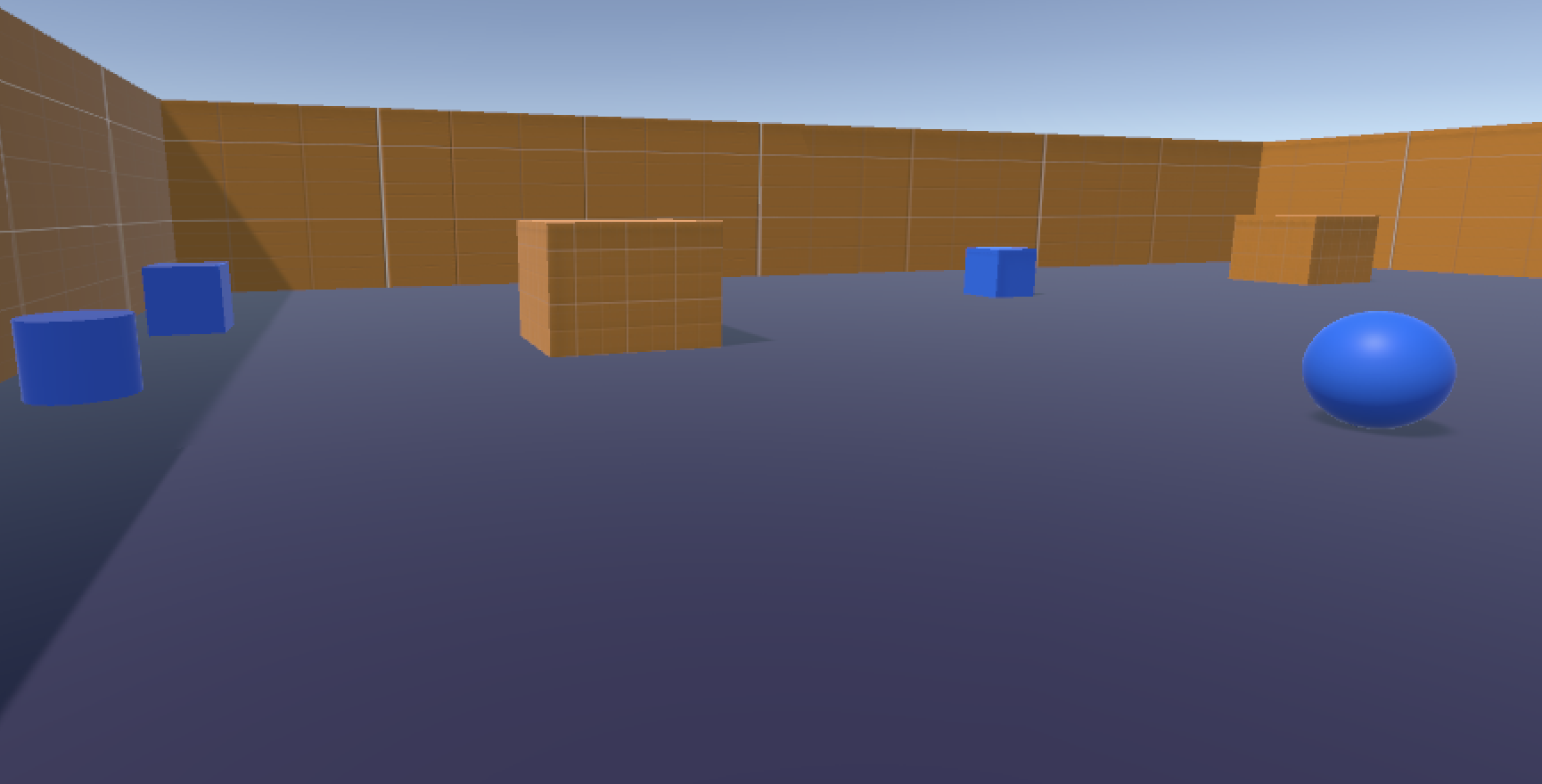}
    \includegraphics[width=.3\textwidth]{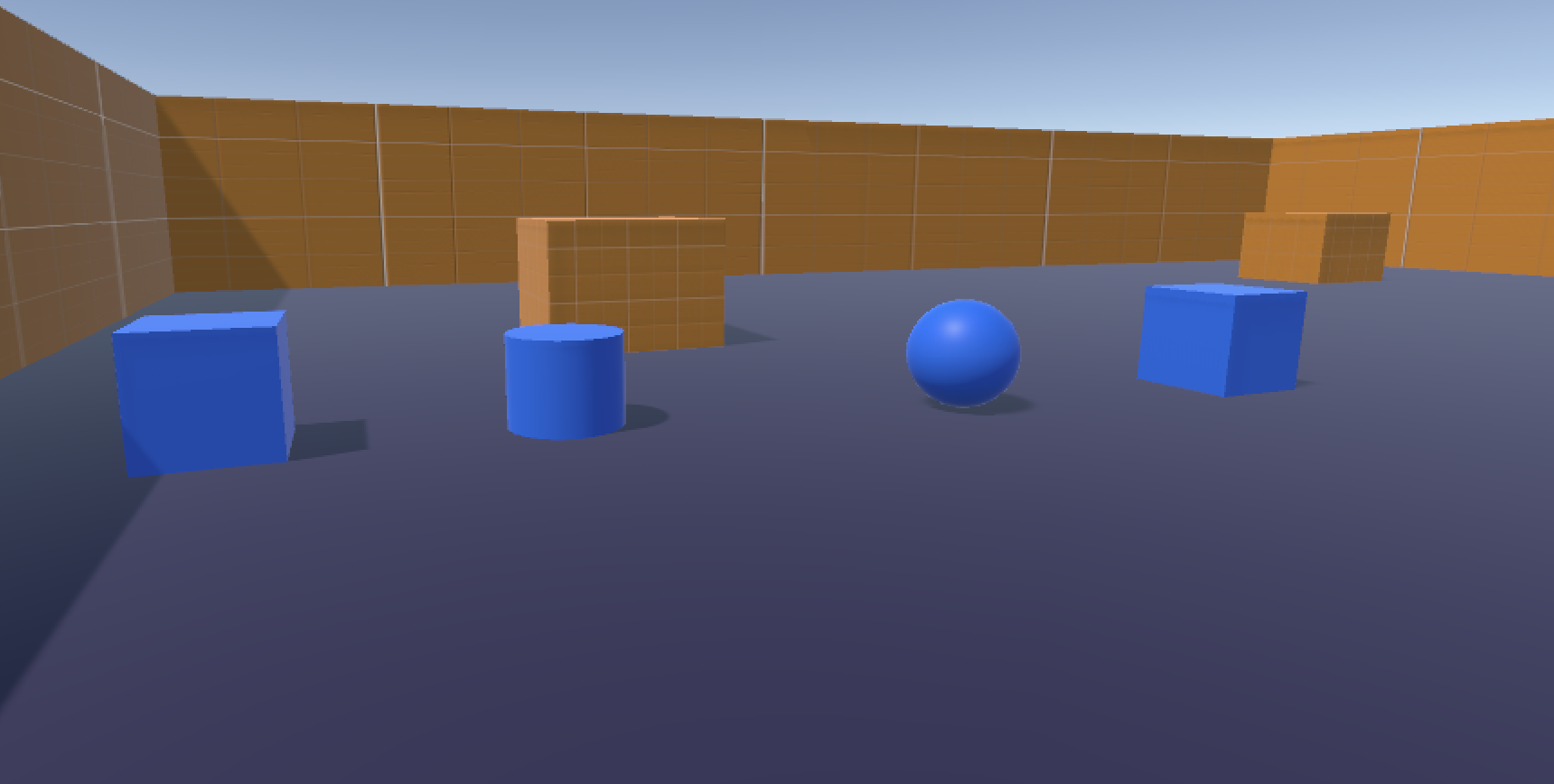}
    \includegraphics[width=.3\textwidth,clip,trim=0 59 0 0]{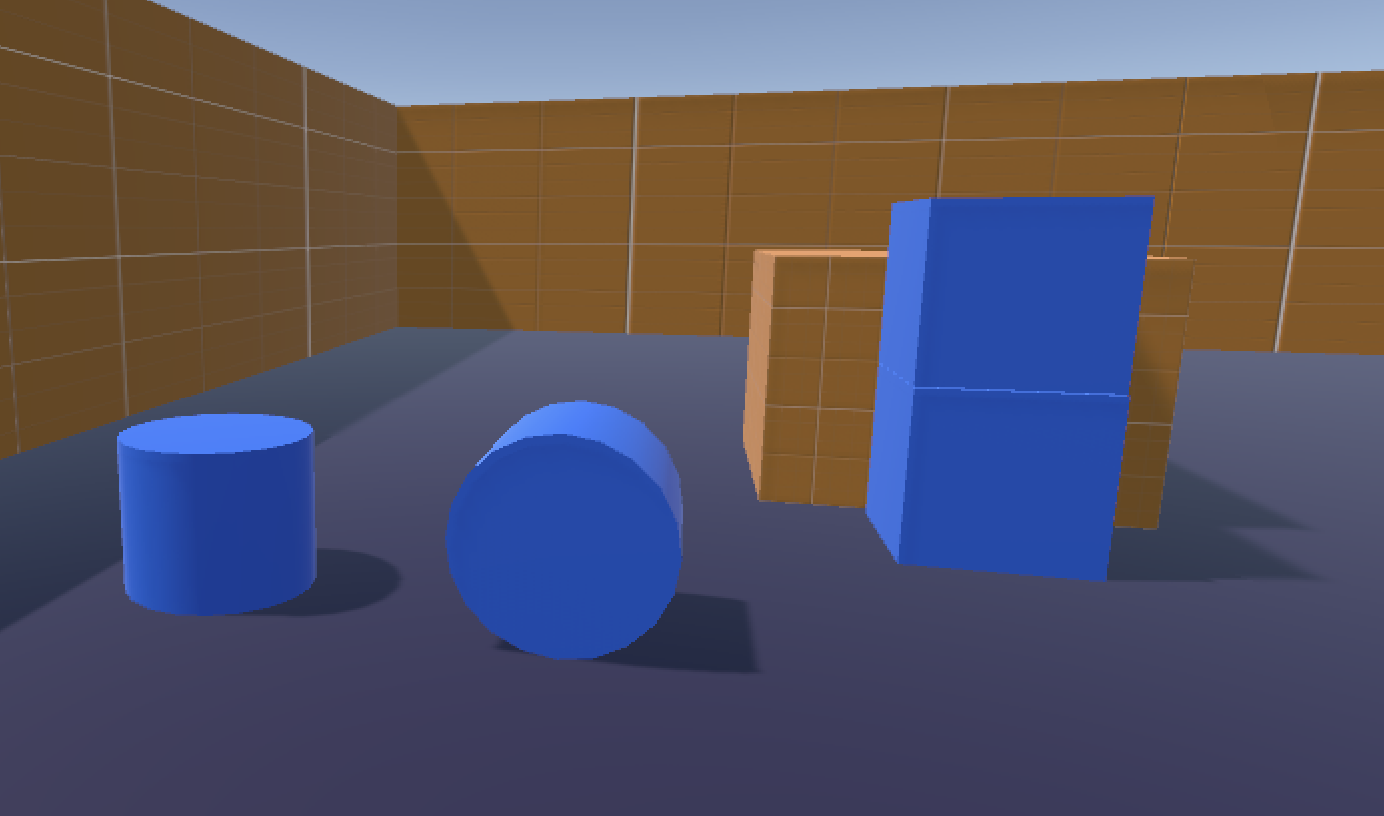}
    \caption{Examples of controls placed on the visual input to LLaVA.}
    \label{fig:controlled-env-scenes}
\end{figure*}

The single-sentence responses from ChatGPT and LLaMA 2 resulted in somewhat better situational descriptions than free-text responses, as did removing the player object distractor in the LLaVA input, so we created a second setting where additional controls could be placed on the environment and visual input.

The second setting involved multimodal input only (hence was only evaluated on LLaVA). We evaluated both free-text responses and responses constrained to be a single sentence. We put a series of controls on the environment to:
\begin{enumerate*}[label=\arabic*)]
    \item Remove the collectible objects so that no distractor objects appeared in the visual input (Fig.~\ref{fig:controlled-env-scenes}[L]);
    \item Place objects directly in the field of view to minimize detection errors due to distance or occlusion (Fig.~\ref{fig:controlled-env-scenes}[C]);
    \item Provide a partial solution to the problem (such as partial object placements and actions taken up until the last step) and ask the model to supply the objects that the final action should be taken with (Fig.~\ref{fig:controlled-env-scenes}[R]).
\end{enumerate*}
In this setting the prompt also included further controls and specificity, such as giving the correct number of objects in the instructions.  Table~\ref{tab:controlled} shows the scores from results in this setting.

\begin{table*}
    \centering
    \small
    \resizebox{\textwidth}{!}{
    \begin{tabular}{lcccccccccccccccc}
    \toprule
    \multicolumn{8}{c}{Free-text} && \multicolumn{8}{c}{1-sentence} \\
    \cmidrule{1-8} \cmidrule{10-17}
    \multicolumn{2}{c}{No distractors} && \multicolumn{2}{c}{All objects in FoV} && \multicolumn{2}{c}{Partial solution} && \multicolumn{2}{c}{No distractors} && \multicolumn{2}{c}{All objects in FoV} && \multicolumn{2}{c}{Partial solution} \\
    \cmidrule{1-2} \cmidrule{4-5} \cmidrule{7-8} \cmidrule{10-11} \cmidrule{13-14} \cmidrule{16-17}
    Stable   & IoU   && Stable   & IoU    && Stable   & IoU && Stable   & IoU   && Stable   & IoU    && Stable   & IoU\\
    \midrule
    1.00$^\dag$         & .75   && .25         & .75    && .67         & 1.00     && .67         & .75   && .67         & .75    && 1.00         & 1.00\\
    \bottomrule
    \end{tabular}}
    \caption{LLaVA scores from controlled environment. $^\dag$While technically none of the objects moved from their final positions after physics was applied, LLaVA's response required placing 2 cubes {\it on top} of the platform and executing a series of jumps with the other objects without stating how they should be placed. This makes this solution invalid.}
    \label{tab:controlled}
\end{table*}

Sample responses, such as ``{\it The two cubes and the sphere can be stacked to reach the top of the platform}'' (free-text response to the \textit{No distractors} condition) and ``{\it The two cubes and the sphere}'' (single-sentence response to the All objects in FoV condition) exhibit many of the same problems as above, and additionally appear to be biased toward trying to use every object in the scene, whether or not that object is actually useful for the task (particularly in the free-text condition). In only one case, where the scene was preset to have completed all but the final step, the choice was between two instances of cylinders in different orientations, and the response was constrained to a single sentence, did the LLaVA model generate an output that correctly solved the task (see Fig.~\ref{fig:controlled-env-scenes}[R], to which LLaVA's response was ``{\it The cylinder on the left should be placed in front of the stack of cubes.}).

\subsection{Visual Grounding with Cross-Attention and BLIP}

A weakness of the LLaVA model is the lack of a cross-modal attention encoder. Instead, LLaVA uses a linear layer to transform visual token embeddings to word embeddings. In contrast, BLIP \cite{li2022blip} uses a more sophisticated cross-modal component to account for the interaction of text and images, which motivates our use of this model. BLIP is a transformer-based multimodal encoder-decoder, including an image-grounded text encoder which uses cross-attention in every other layer to ground individual words of a text caption to regions of an image. Unlike LLaMA 2/LLaVA/ChatGPT, BLIP does not output long free-text responses to open-ended questions, so it is unsuitable for generating outputs that can be evaluated by operationalizing them in the simulation.  However, it is jointly pretrained with three loss functions: image-text contrastive learning, image-text matching, and image-conditioned language modeling, which provides a more sophisticated way of seeing where visual attention is being applied when conditioned on language {\it input} than LLaVA's linear projection. In this section we use BLIP as a representative example of an architecture that models the interaction of images and language through cross-attention to examine if and where cross-attention would better capture correct portions of the image containing terms key to physical reasoning problems such as ours. We examine this with a  curated set of inputs, and find that in many cases such a model still fails to ground properties/concepts such as flatness/roundness given an image.  Examples follow.\footnote{The figures omit visual grounding of articles like ``the'' or ``a.''}

\begin{figure*}[h!]
    \centering
    \includegraphics[width=.95\textwidth]{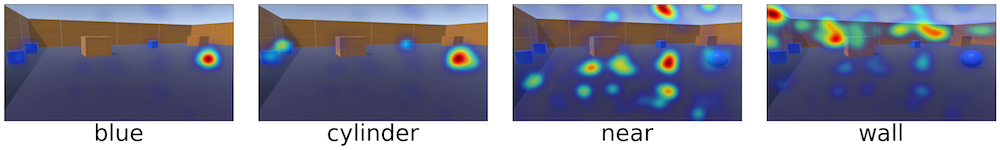}
    \includegraphics[width=.95\textwidth]{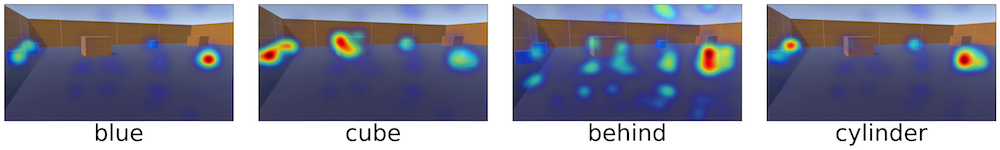}
    \caption{Per-word BLIP visual grounding for ``{\it blue cylinder near wall}'' (top) and ``{\it blue cube behind cylinder}'' (bottom).}
    \label{fig:blip-cube-cylinder}
\end{figure*}

In Fig.~\ref{fig:blip-cube-cylinder}, the cylinder is on the far left and the sphere is on the far right of the source image.  When the word ``cylinder'' is given, the most heavily weighted object ends up actually being the sphere on the right. This may point to a bias toward round objects like spheres in the object localization components, and perspective distortion effects on the appearance of the cylinder may make it appear more like another object type, such as a cube.  When the word ``cube'' is given, a significant portion of the weight falls on two objects: the cubical platform that is part of the environment, and the cylinder. 

\begin{figure*}[h!]
    \centering
    \includegraphics[width=.45\textwidth]{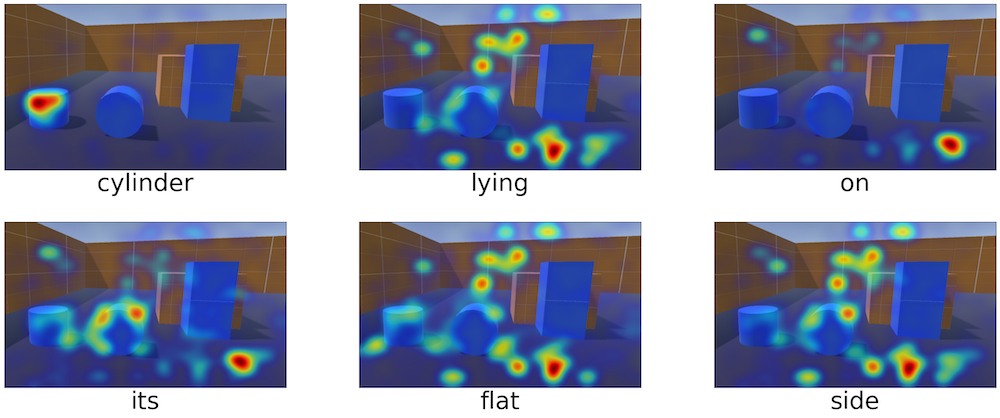}
    \hspace{.05\textwidth}
    \includegraphics[width=.45\textwidth]{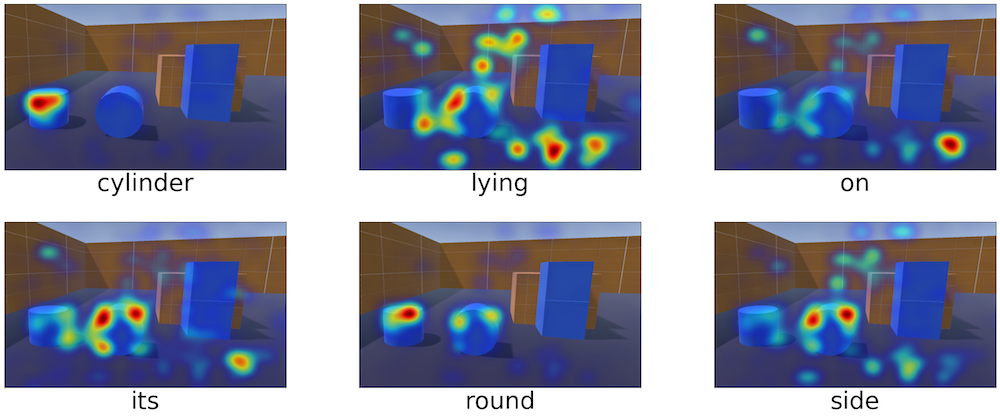}
    \caption{Per-word BLIP visual grounding for ``{\it cylinder lying on its flat side}'' [L] and ``{\it cylinder lying on its round side}'' [R].}
    \label{fig:blip-flat-round}
\end{figure*}

\begin{figure}[h!]
    \centering
    \includegraphics[width=.45\textwidth]{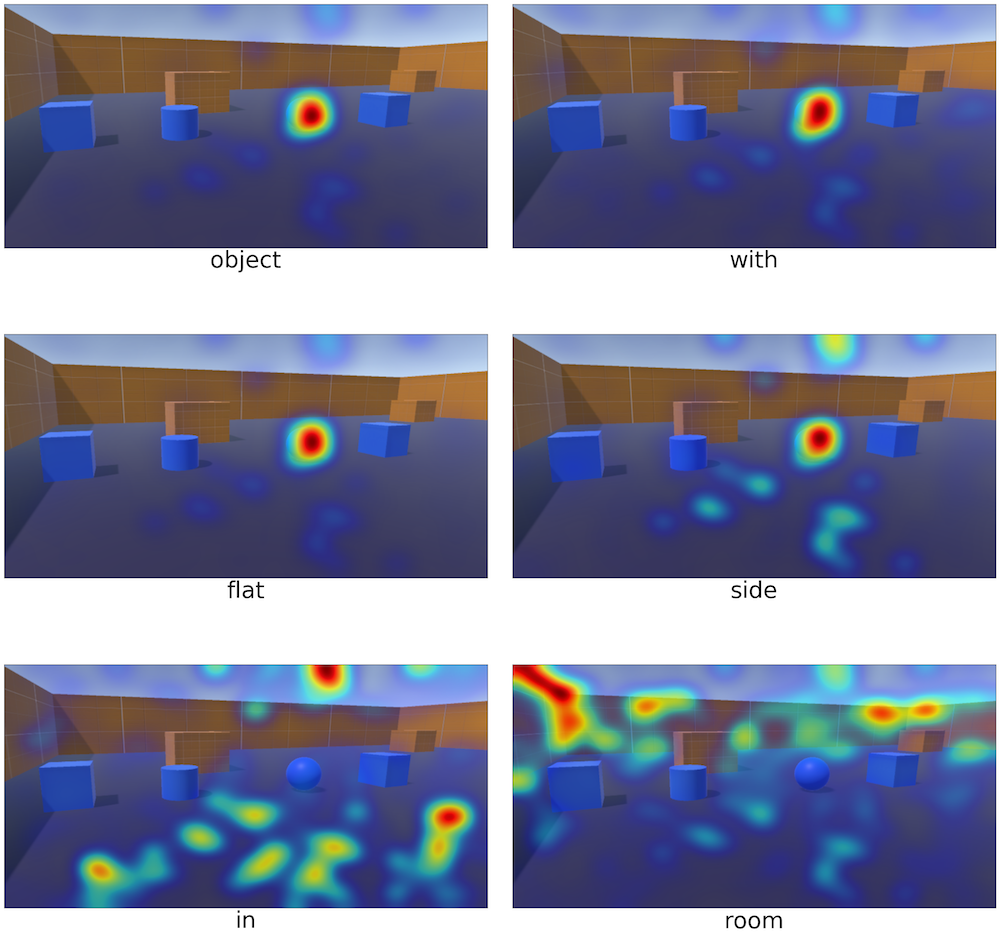} \\
    \vspace{10mm}
    \includegraphics[width=.45\textwidth]{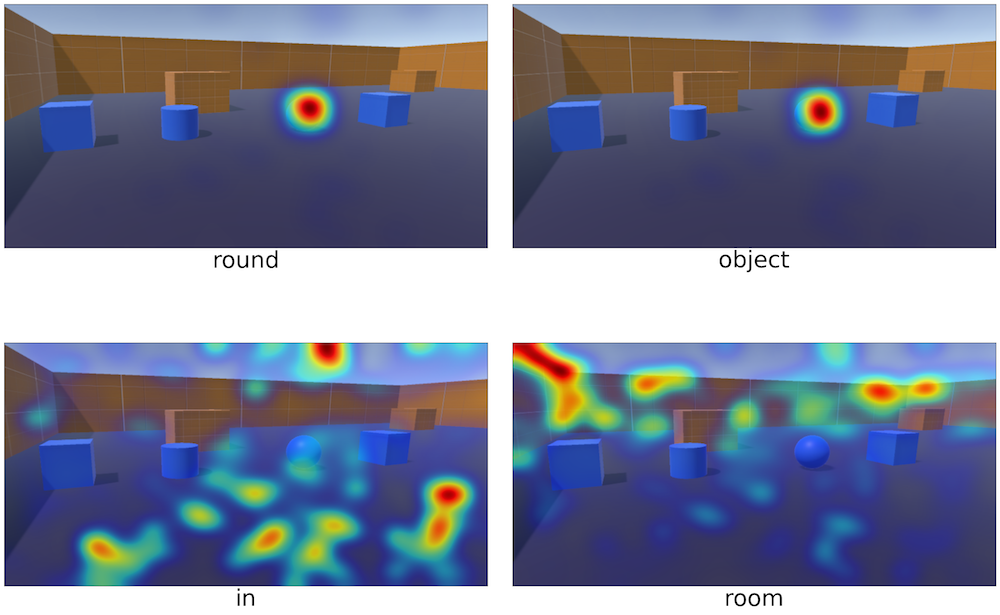}
    \caption{Per-word BLIP visual grounding for ``{\it object with flat side in room}'' (top) and ``{\it round object in room}'' (bottom).}
    \label{fig:blip-flat-round-2}
\end{figure}

\begin{figure}[h!]
    \centering
    \includegraphics[width=.45\textwidth]{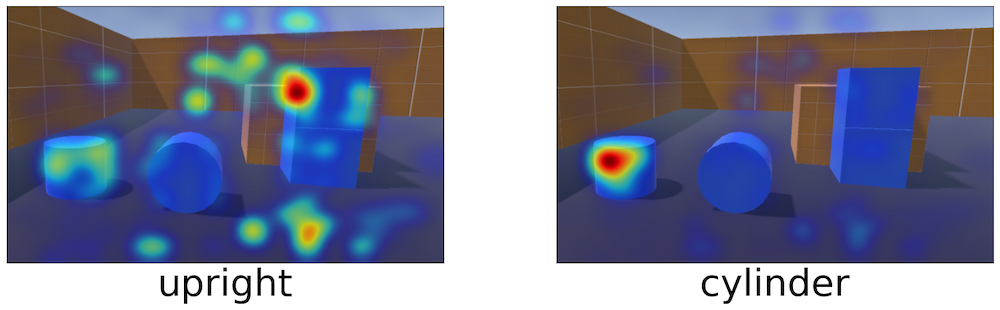}
    \caption{Per-word BLIP visual grounding for ``{\it upright cylinder}''.}
    \label{fig:blip-upright-cylinder}
\end{figure}

In Fig.~\ref{fig:blip-flat-round} we see that despite there being two cylinders in the scene, the word ``cylinder'' is strongly grounded to the upright cylinder (resting on its flat side), even when the text prompt mentions the cylinder on its {\it round} side. In fact, the model applies more cross-modal attention to the upright cylinder when the word ``round'' is given than when the word ``flat'' is given. Likewise, in Fig.~\ref{fig:blip-upright-cylinder}, the highlighted region falls only on the upright cylinder and not the other cylinder. This is likely because cylinders lying on the round side are sparsely or not present in image/text pairs during pretraining. As \citet{barbu2019objectnet} observed, common image datasets often contain biases toward objects in canonical positions.  For cylindrical objects, this includes things like soda cans or glasses in the canonical upright orientation.

To examine the extent to which concept words such as \textit{flat} and \textit{round} are visually grounded in a zero-shot manner, we consider various captions (for different images and points of view in our environment) containing these words. Fig.~\ref{fig:blip-flat-round-2} further illustrates that concepts such as ``round'' and ``flat,'' critical to our scenario, are not really grounded in this model. The sphere is highlighted for both, indicating that ``round'' is successfully localized to the sphere likely because of a significant presence in the training data. Despite the contrastive loss used for training BLIP, the antonym ``flat'' is not localized to a non-round object.


Taken together, these results suggest that even a pretrained vision-language model that contains even a sophisticated cross-attention architecture suffers from problems in grounding descriptive terms to the correct instances of objects, unless supplemented with other, non-linguistic, non-visual information.

\section{Exploiting Plan Failure}

\begin{figure*}
    \centering
    \includegraphics[width=.3\textwidth,clip,trim=150px 50px 150px 50px]{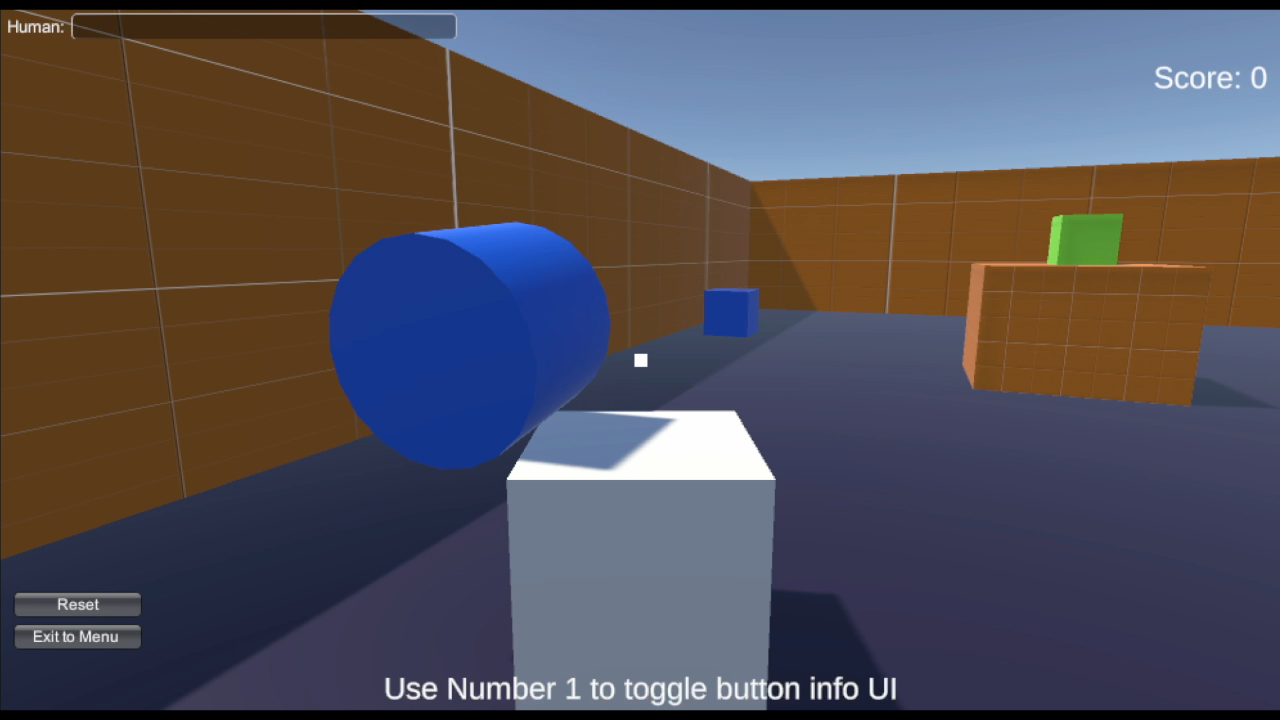}
    \includegraphics[width=.3\textwidth,clip,trim=150px 50px 150px 50px]{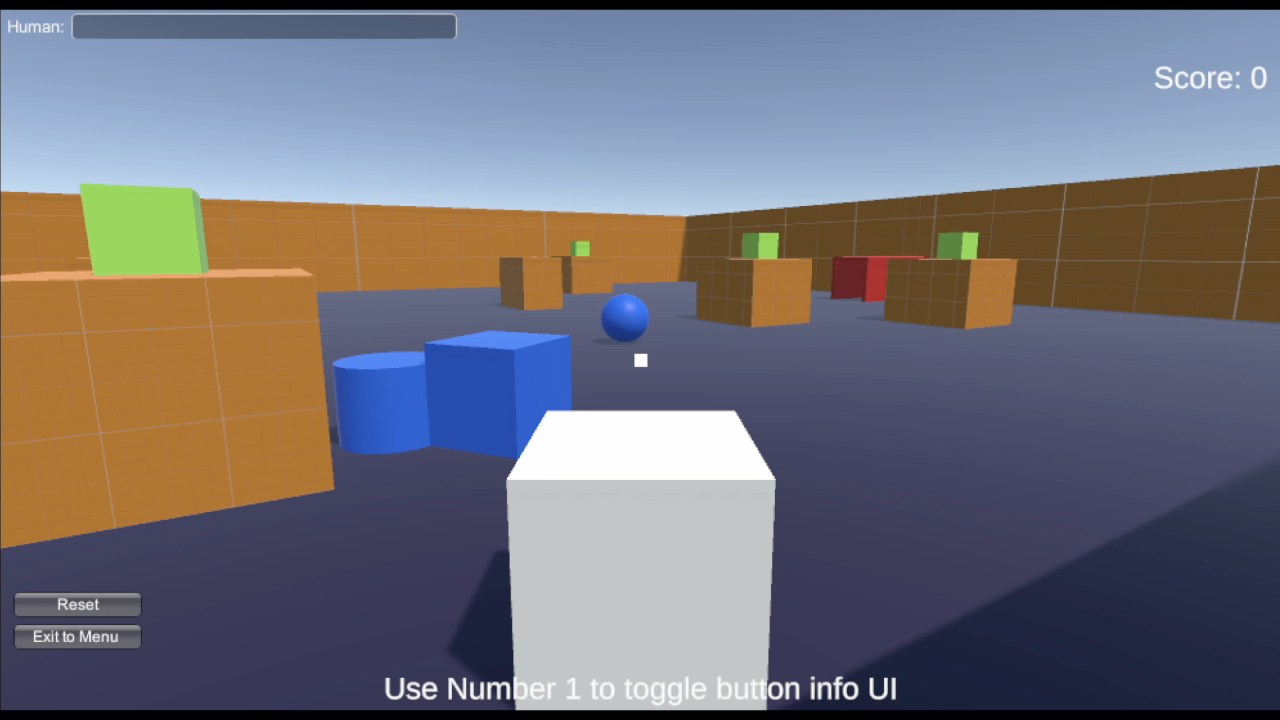}
    \includegraphics[width=.3\textwidth,clip,trim=150px 50px 150px 50px]{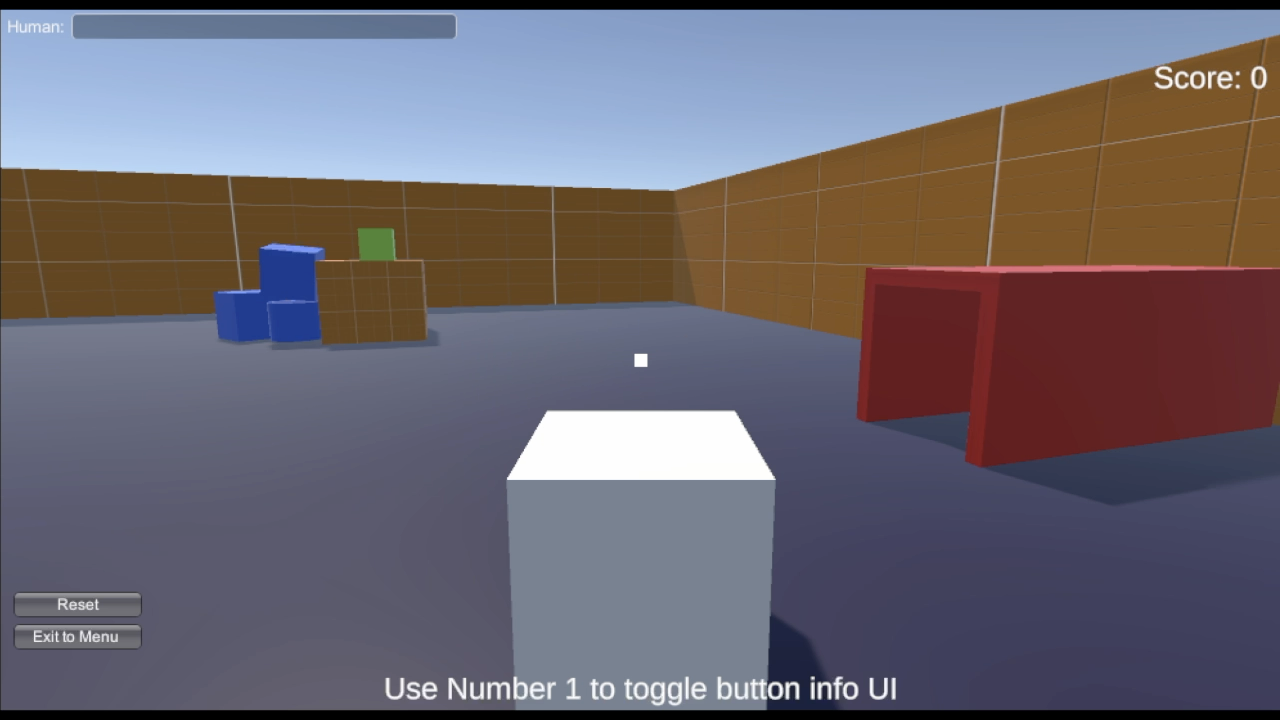}
    \caption{Agent executing plan with our method. L: Agent samples cylinder. C: Agent places cylinder flat side down, then samples cube and places it beside cylinder. R: Agent samples cube and places it on top of cylinder.}
    \label{fig:sampling}
\end{figure*}

Given the results above, it is clear that both text and multimodal language models struggle with correctly modeling the environmental dynamics even in simple situations like ours. This appears to primarily be a shortcoming of using text alone or only text and images as the model inputs, when other modalities besides these are truly informative about object properties and behaviors.

As humans develop object concept representations, they are also learning to individuate objects from the perceptual flow not just based on visual features but also based on experience that includes interacting with them in real time~\cite{spelke1985perception,spelke1990principles,baillargeon1987object,spelke1989object}. We therefore developed a procedure through which an agent embodied in a scene performs a semi-self-guided exploration of objects it encounters and uses that to solve the problem by linking object classifications in the scene to a background knowledge base.

Our knowledge base in this case is a library of {\it voxemes} in the VoxML modeling language~\cite{pustejovsky2016voxml}. VoxML models object relevant properties like symmetry, habitats (conditioning environments, as in \citet{pustejovsky2013dynamic}), and affordances, which makes it useful for specifying realistic object behavior in a continuous simulation environment, but our method is in principle friendly to an arbitrary background knowledge base as in~\citet{nirenburg2023hybrid}.

Within the simulation, as objects are interacted with and move, they leave trajectory traces through the environment, governed by the underlying VoxML semantics in interaction with the physics engine. This information includes position, rotation, and movement in response to various actions. For instance, this would take a cylinder's major axis of symmetry from VoxML (the $Y$ axis), and make the cylinder roll along that axis if rolling is an available affordance given the cylinder's current habitat.

In our procedure, we initially naively follow the step-by-step plan generated by an LLM.  At some point, proceeding further in the specified plan becomes impossible. For example, if trying to build the configurations shown in Fig.~\ref{fig:ex}, the agent cannot balance or stack another object on the sphere, which triggers an exploration process.

As our principle concerns in this environment involve successfully stacking objects so they can either support other objects or an agent, we build our exploration strategy on a previous model from \citet{ghaffari2023grounding}. This underlying model uses data gathered in two-object stacking task to classify a set of objects (the same 3 object types as used above, as well as 6 others, which display contrasts between the flatness and roundness, depending on the orientation of the objects). It uses similarity learning to create two high-level clusters in the model's latent space: one that contains the \textit{flat} objects, or objects that largely remain stable when stacked on a flat surface, and the \textit{round} objects, which do not.
Importantly, the model is trained only over objects that are flat on all sides (cube, pyramid, etc.), or entirely round (sphere, egg, etc.). Objects like the cylinder, which display both qualities depending on orientation, are not included in training. This means that the process of grounding the cylinder to the appropriate region involves determining the orientation in which that object is considered \textit{flat} vs. \textit{round}, rather than simply classifying the object.

When the failure of the initial LLM-generated plan triggers exploration, the agent traverses the environment, selects objects, and attempts to stack each one on top of itself. Because the agent is represented as a cube, the exploration process replicates the original stacking task the model was trained on. The agent has a goal to seek out objects that can be grounded to the \textit{flat} region in its underlying model. 

\begin{figure}[h!]
    \centering
    \includegraphics[width=.95\columnwidth]{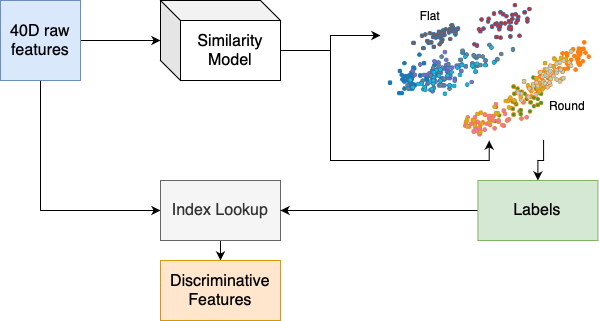}
    \caption{Schematic overview of exploration procedure.}
    \label{fig:diagram}
\end{figure}

As the cylinder is unseen by the underlying model, the agent makes its judgments through analogizing the behavior of samples of the new object to samples of previously-encountered objects using vector similarity in the latent space. It then indexes the conditions under which those behaviors occur, which links the raw trajectory features to the embedding vector and its nearest neighbors.  A schematic overview of this procedure is shown in Fig.~\ref{fig:diagram}.

In our method, the agent samples an object, determines the configuration, or habitat, the object requires to satisfy the stackability property, and places it, resulting in a sequence of events like that shown in Fig.~\ref{fig:sampling}.  The agent first samples the cylinder and determines how to place it so it can participate in a successful stacking relation (flat side down).  It then samples a cube and places it beside the cylinder.  It then samples another cube and places it on top of the previously-placed cylinder, creating an approximation of a 2-step staircase that can be used to climb the platform.

\section{A Proposal for Grounding Language Models to Physical Laws}

We have demonstrated that current LLMs struggle with tasks of this nature, and have proposed a method to use exploration of objects under interaction to achieve a successful solution in this class of problems.  The technical challenge is to retain LLMs' open-domain, grammatical generation capabilities while accounting for environmental dynamics and object properties {\it in situ}.

We frame this as a knowledge distillation problem \cite{hinton2015distilling} with the added challenge of needing to transfer knowledge into an LLM from a model specialized to object classification over images and trajectory data. Even a relatively small LLM like LLaVA-7B is still significantly larger than a candidate object classification model like those we have used in this line of research so far, and a standard soft logit distribution over the object class labels is not likely to provide sufficient information for a generative LLM with thousands of potential token outputs.

In addition, there is a substantial distribution mismatch between the autoregressive language model and the models in which we are encoding the information we want to source, such as relations between objects, their properties, and their afforded behaviors. This information has been bootstrapped through a knowledge base or modeling language like VoxML, and needs to be first vectorized into a subsymbolic form, and those vectors need to be aligned with the distribution of the generative LLM.

To address this challenge, we propose leveraging the information that can be directly extracted from a simulation environment like ours.  We know the location and extents of different objects in space, and the camera position from which images are captured.  All these are expressed in Cartesian coordinates and a quaternion for rotations, compressed into a single 4$\times$4 transformation matrix.  Therefore, we propose projecting the object locations from 3D world space into pixel space to target patches in the images where we know attention should be paid if the correct object is to be extracted from the image. The spatial trajectory data, their localization in pixel space (e.g., as bounding boxes), along with the images, would be passed into a transformer encoder. Self-attention would be trained to detect objects in the image, with an additional object localization signal from the bounding box with the object label attached.



Attention to the correct region of the image features should also condition tokens in the outputs that describe object-relevant properties or actions. In \citet{ghaffari2023grounding} we already showed that by grounding object terms from a language model to objects from a trajectory-based classifier, we get information about related terms ``for free'' (e.g., grounding terms for flat objects to the flat cluster also grounded terms like ``stack'').


We propose to encode the raw object movement and/or visual features in a self-supervised fashion through an attention encoder, to tease out correlations in object features relevant to object identification. For example, the major axis of a cylinder, i.e., as encoded in VoxML, is a strong correlate to its ``stackability'' or flat surface in one orientation, and its lack thereof in another. Since this information is implicitly encoded in the object trajectory features from the simulation, we want to enable the language model to learn correlations between them and task-relevant tokens.

Supervision can then be sourced from attention heads over the object encodings to those of the language model. Given the previously-uncovered correlations between object and behavior terms in language models, an attention loss (Eq.~\ref{eq:att}, for up to $h$ attention heads, where $L$ denotes the language model and $O$ denotes the object model) should optimize the model to apply attention from a cylinder representation to related action and property tokens, like ``stack'' if the cylinder is vertical, or ``roll'' if it is horizontal. $a_{i}$ denotes the \textit{i-th} attention head. $\boldsymbol{Obj}$ denotes the representation obtained by language model ($L$) or object model ($O$). This loss function minimizes the Euclidean distance between attention from the object and language models.

\begin{equation}
    \mathcal{L}_{\text {att}} = \sum_{i=1}^h\|\boldsymbol{Obj}_{a_i}^L-\boldsymbol{Obj}_{a_i}^O\|_2^2
    \label{eq:att}
\end{equation}

An embedding loss (Eq.~\ref{eq:emb}) then would minimize the Euclidean distance between object embeddings drawn from the final hidden state ($s$) of the object classifier and those of object denoting tokens. Because the two models are trained over different data with different initializations and regimes, the resulting embedding spaces are not directly comparable, so a learned linear matrix, denoted $\boldsymbol{W}_{V\rightarrow L}$, projects the visual embeddings into the language model's space.

\begin{equation}
    \mathcal{L}_{\text {emb}} = \|\boldsymbol{Obj}_{s}^L - \boldsymbol{Obj}_{s}^V\boldsymbol{W}_{V\rightarrow{}{L}}\|_2^2
    \label{eq:emb}
\end{equation}

\begin{figure}[h!]
    \centering
    \includegraphics[width=.95\columnwidth]{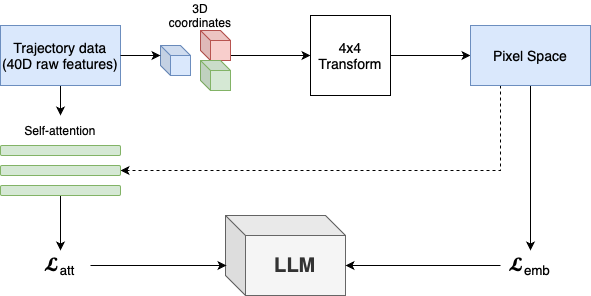}
    \caption{Proposed architecture for distilling object property and motion information to LLM representations.}
    \label{fig:proposal}
\end{figure}

Fig.~\ref{fig:proposal} shows a design schematic of the proposed model. 

Within the LLM, a contrastive method {\it a la} \citet{yang2023rlcd} provides a method for generating both positive and negative samples of LLM responses for scoring and training a preference model, which addresses a profound challenge in standard reinforcement learning from human feedback (RLHF) approaches~\cite{christiano2017deep}: sourcing a sufficient distribution of scored responses is expensive and time consuming. However, here we are only concerning ourselves with generating physically sensical and plausible responses that account for object properties and the operation of environmental dynamics over them. Therefore, we propose to source the scoring directly from the simulation itself. For example, if we are concerned with optimizing the LLM toward generating correct responses of physical reasoning about properties relevant to object stacking, then a ``good" response to prompts of this nature would be those that, when operationalized, generate stable configurations. ``Bad" responses would be those that either cannot be logically satisfied (e.g., due to placing multiple objects in the same place), or results in unstable configurations.  This would enable us to rapidly source a preference signal in a task-focused fashion.



The preference model is then trained using the sum of a contrastive loss, the attention loss, and the embedding loss with weighting terms $\Lambda$ applied to each function and tuned over a validation set to minimize the average loss. Eq.~\ref{eq:comb} provides the combined loss function.

{\begin{equation}
    \mathcal{L} = \lambda_1\mathcal{L}_{\text {cont}}+\lambda_2\mathcal{L}_{\text {att}}+\lambda_3\mathcal{L}_{\text {emb}}
    \label{eq:comb}
\end{equation}

The outputs of the preference model flow into the RL training of the LLM, using existing methods such as proximal policy optimization (PPO).

\section{Conclusion}

Despite the problems in LLMs' reasoning that we have demonstrated, and that others have also studied in different domains, the way that LLMs are discussed in both scientific literature and popular media \cite{shanahan2022talking} creates an impression of human-like abilities. Sometimes these apparent abilities do manifest in certain domains and conditions, but this should not be taken as an indicator of general human-like reasoning abilities, especially in the very physically- and situationally-grounded type of problems we examine.

In this paper we presented evidence of the difficulty LLMs have in physical reasoning problems, by mediating the LLMs's solutions through a simulation environment to examine the effects of physics and environmental dynamics on the presented solutions. Problems manifested particularly when executing a multi-step plan where completing future steps depends on successfully exploiting the relationships created between object during past steps. We focused on multimodal models, specifically LLaVA due to its open-weight nature, and compared model outputs from this and text-only models, in different conditions, including allowing free-form or restricted responses, and putting controls on the visual input from the environment to clearly select for the information needed to solve the problem without distractors. Only in one condition were we able to coax the correct solution from the LLM: when all but the last step of the presented problem were already given, and the LLM simply had to choose the object to be used in the final step. The results point to a weakness in LLMs at reasoning \textit{causally}, such as successfully predicting what will happen to an object configuration due to the application of consistent environmental physics after it is created. We also investigated BLIP and identified cases where it still cannot ground concepts and properties inherent to the objects in our environment in a zero-shot manner despite large-scale image/text pretraining.

This led to the second part of our contribution: an alternative method for determining the right object to be used for the task through interaction and exploration, and a proposal to use the information extracted from such explorations in the LLM to direct outputs toward better (in this case, more physically feasible) solutions by distilling grounding signals from the simulation environment back into the model. This amounts to a process of making what is implicit in linguistic input more explicit~\cite{krishnaswamy2017monte}, and opens up many more avenues in future work toward grounding LLMs in a realistic understanding of causality and natural laws.

\bibliography{aaai24}

\end{document}